\documentclass{article}

\usepackage{PRIMEarxiv}

\usepackage[utf8]{inputenc} % allow utf-8 input
\usepackage[T1]{fontenc}    % use 8-bit T1 fonts
\usepackage[numbers]{natbib}
\usepackage{hyperref}       % hyperlinks
\usepackage{url}            % simple URL typesetting
\usepackage{booktabs}       % professional-quality tables
\usepackage{amsfonts}       % blackboard math symbols
\usepackage{nicefrac}       % compact symbols for 1/2, etc.
\usepackage{microtype}      % microtypography
\usepackage{lipsum}
\usepackage{fancyhdr}       % header
\usepackage{graphicx}       % graphics
\usepackage{amssymb} 
\usepackage{amsmath} % for \left\lVert and \right\rVert

\graphicspath{{media/}}     % organize your images and other figures under media/ folder

\title{Graph Neural Networks for Structural Displacement Prediction}

\author{
  Hung-Fu Chang \\
  R. B. Annis School of Engineering \\
  University of Indianapolis \\
  Indianapolis\\
  United States\\
  \texttt{hchang@uindy.edu} \\
   \And
  Tzu-Kang Lin, Yung-Li Cheng \\
  Department of Civil Engineering \\
  National Yang Ming Chiao Tung University \\
  Hsinchu \\
  Taiwan \\
  \texttt{\{tklin, leoncheng.en13\}@nycu.edu.tw} \\
}

\begin{document}
\maketitle

\begin{abstract}
Accurate prediction of structural displacements under external loading is fundamental to structural health monitoring and seismic safety assessment. Although the finite element method (FEM) remains the prevailing approach because of its high accuracy, its considerable computational cost restricts its suitability for real-time monitoring applications. To address this limitation, this study proposes a data-driven framework based on Graph Neural Networks (GNNs), in which structural systems are represented as graphs with joints modeled as nodes and structural members as edges. By incorporating both geometric and mechanical properties into the graph representation, the proposed model learns the relationship between applied loads and structural responses directly from simulated data. A synthetic dataset was generated from a two-story frame structure using ANSYS, and both a conventional Neural Network (NN) and a GNN were trained for comparison. The results show that the proposed GNN framework predicts displacements and rotations with high accuracy and outperforms the NN model, demonstrating its potential as a fast and efficient alternative to traditional FEM-based analysis.
\end{abstract}

% keywords can be removed

Accurate prediction of structural displacements is a fundamental problem in structural engineering because displacement response is directly related to serviceability, damage evaluation, and the overall safety of structural systems subjected to external loads. In practice, reliable displacement estimates are essential for performance assessment under dynamic and quasi-static loading conditions, such as earthquakes, wind, and other operational demands. They are also critical for structural health monitoring, condition assessment, and the development of rapid decision-support systems for infrastructure management.

For decades, structural response prediction has been dominated by analytical formulations and numerical simulation techniques, most notably the finite element method (FEM) \cite{hughes_2000_the}. FEM-based analysis follows a bottom-up strategy in which a structure or an object is discretized into a large amount of elements, local stress and strain are computed at the element level, and the global displacement response is subsequently assembled from these local interactions. Because of FEM method’s physical foundation and ability to model complex geometries, boundary conditions, and various material behaviors, It has become the standard tool for structural analysis and simulation. 

While FEM is robust and physically interpretable, high accuracy simulations typically demand substantial preprocessing effort, repeated numerical solution procedures, and considerable computational resources \cite{shivadityamedurivenkata_2022_graph}. This challenge becomes even more pronounced for structures with intricate geometries or multiple constituent materials, where the complexity of the system significantly increases the cost of modeling and analysis. Consequently, their application becomes limited in scenarios that require rapid evaluation, repeated analysis over many loading cases, or near-real-time response prediction.

Recent advances in machine learning have introduced a new paradigm for structural analysis, shifting the emphasis from purely physics-based computation to data-driven modeling \cite{maurizi_2022_predicting, pfaff_2021_learning, wrth_2024_physicsinformed}. Unlike conventional bottom-up approaches, machine learning methods consider the structure as an integrated system and learn the relationship between structural inputs, such as external loads, and structural responses, such as displacements, directly from data. In this framework, the structure is treated as an input–output system, without explicitly resolving the detailed behavior of individual materials at every stage of analysis. Once trained, the model can predict structural displacements under new loading conditions without repeatedly solving the governing equations, thereby significantly reducing computational cost. Moreover, such approaches offer strong potential for practical implementation in real-world applications, particularly when sensor systems are available to provide measurements of loads, displacements, and deformations. These advantages make machine learning especially attractive for applications requiring rapid prediction, scalability, and efficient repeated inference.

Graph neural networks (GNNs), initially introduced in 2005 \cite{gori_2005_a}, are designed to process data that can naturally be represented as graphs. They address the limitations of traditional machine learning methods in handling topological relationships and data structures \cite{zhou_2020_graph, zhao_2024_a}. Therefore, GNNs are particularly effective for problems in which the interactions among connected entities can be expressed in a graph. This characteristic makes GNNs particularly attractive for structural analysis, since a structural system can be naturally represented as a graph, where joints are treated as nodes and beams or columns as edges. Such a representation preserves the physical connectivity of the structure and allows the model to capture the influence of member-to-member interactions on the global response.

Motivated by this perspective, this study investigates the application of graph neural networks (GNNs) to structural displacement prediction, a topic that remains relatively underexplored in existing literature. The main objective is to assess whether a GNN can achieve accurate and computationally efficient prediction of structural response in comparison with other machine learning models. To facilitate model training and evaluation, a synthetic data generation strategy is employed to produce training and evaluation datasets. Specifically, the nodal displacements and rotations of a two-story structural frame subjected to external loading are investigated within machine learning models. In this way, the study provides insights of using machine learning methods for structural analysis as an initial contribution to the growing body of research interest on machine learning-based structural analysis. We also contribute a dataset that can be used in the future machine learning based studies. Both point to the potential of such approaches for future applications in structural health monitoring and simulation.

\section{Literature Review}\label{sec2}
Existing applications of Graph Neural Networks (GNNs) in civil and structural engineering can generally be grouped into two main directions. The first focuses on mesh-based simulation and surrogate modeling for structural analysis. A representative study in this area is MeshGraphNets proposed by \cite{pfaff_2021_learning}, a general-purpose framework for mesh-based simulation that employs message-passing GNNs on irregular meshes to model physical systems efficiently. Their work addresses a solution for the high computational cost due to the high-dimensional numerical simulations. Align with this foundation, \cite{wrth_2024_physicsinformed} proposed Physics-Informed MeshGraphNets (PI-MGNs), which integrate MeshGraphNet with physics-informed training through FEM/PDE constraints for non-stationary and nonlinear simulations on arbitrary meshes, further improving computational efficiency while preserving physical consistency. Similarly, \cite{shivadityamedurivenkata_2022_graph} developed a GNN-based model as an alternative to FEM in order to reduce the time required for simulation. More recently, \cite{iparraguirre_2026_meshgraphnettransformer} extended MeshGraphNet by coupling it with a physics-attention Transformer to better capture long-range interactions, thereby addressing the limited efficiency of iterative message passing on large, high-resolution meshes. This improvement is particularly relevant for large-scale civil structures, where distant structural coupling may play an important role.
The second major application of GNNs in this field concerns structural damage detection and health monitoring. \cite{wijethunga_2025_robust} introduced a dual-graph framework in which one graph represents sensor topology and another capture feature correlations, combined with a CNN-based denoising and compression module for vibration signals. Their method demonstrated strong accuracy and robustness on benchmark datasets while addressing practical constraints in structural health monitoring (SHM). \cite{dang_2022_structural} proposed a graph representation that incorporates both structural geometry and measured vibration data, and applied Graph Convolutional Networks to exploit spatial correlations among sensor locations for damage identification. ~\cite{kim_2025_nearrealtime} developed a dynamic GNN (DynGNN) in 2024 for near-real-time damage identification, improving upon conventional approaches based on static graph structures. Their validation on test data from a steel truss bridge under vehicle loading suggested promising applicability to real-world scenarios. \cite{li_2026_dataphysics} proposed a data-physics fusion framework using spatiotemporal graph signal analysis for bridge damage diagnosis, with an emphasis on extracting coupled temporal, frequency, and spatial features to improve damage localization. In a related study, \cite{liu_2026_lgstagnn} proposed a Local-Global Spatiotemporal Attention (LGSTA) GNN for bridge damage detection, highlighting the need to capture both local sensor-neighborhood patterns and global structural context in vibration-based damage assessment.
Although these studies demonstrate the growing potential of GNNs in mesh-based simulation and structural damage identification, their application to structural displacement prediction remains relatively limited. This gap suggests the need for further investigation into the use of GNNs as an efficient and physically meaningful approach for structural analysis.

\section{Approach}\label{sec3}
\begin{figure*}[t]
\centerline{
\includegraphics[width=0.77\textwidth]{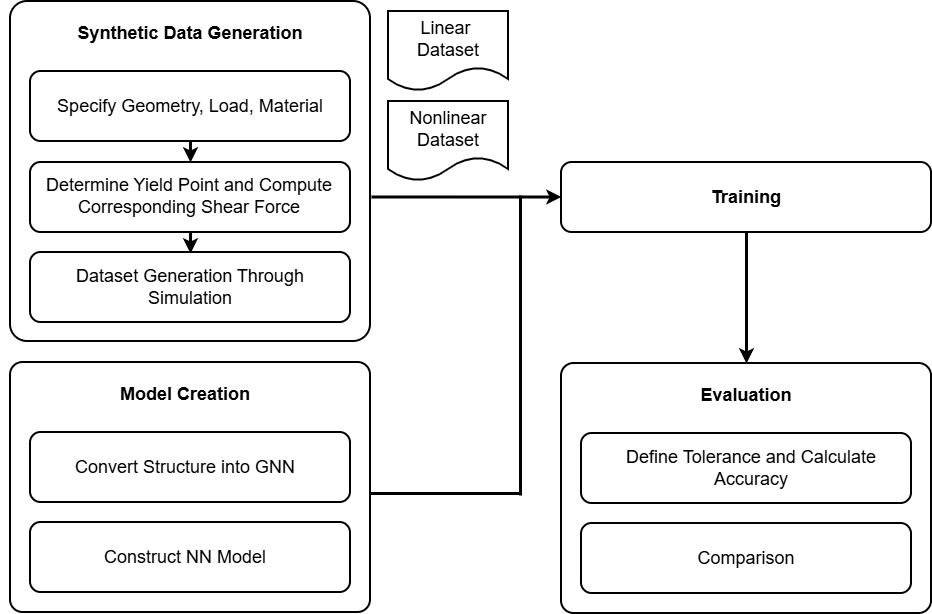}
}
\caption{The overall approach of this study.\label{fig1}}
\end{figure*}

Our method consists of four major phases: synthetic data generation, model creation, model training and result evaluation (see ~\autoref{fig1}). The synthetic data generation produces linear and non-linear datasets. These two datasets were determined by the yield point of the specified material. The yield point can help us to compute the yield shear force by using the Portal Method. After yield shear force can let us to determine the material is under linear or nonlinear state. After using incremental loads on ANSYS simulation, we can get the nodal movement at x, y direction and rotation at z axis. The displacement data will be arranged in the dataset.  

% figure1

To investigate GNN’s performance, we not only examine the accuracy of the training result but also compare with NN model’s training result to know if GNN is a much better choice. The GNN and NN models are formed differently because we form the directed graph by considering the structure properties.

\subsection{Graph Representation and Features}
\begin{figure*}[t]
\centerline{
\includegraphics[width=1\textwidth]{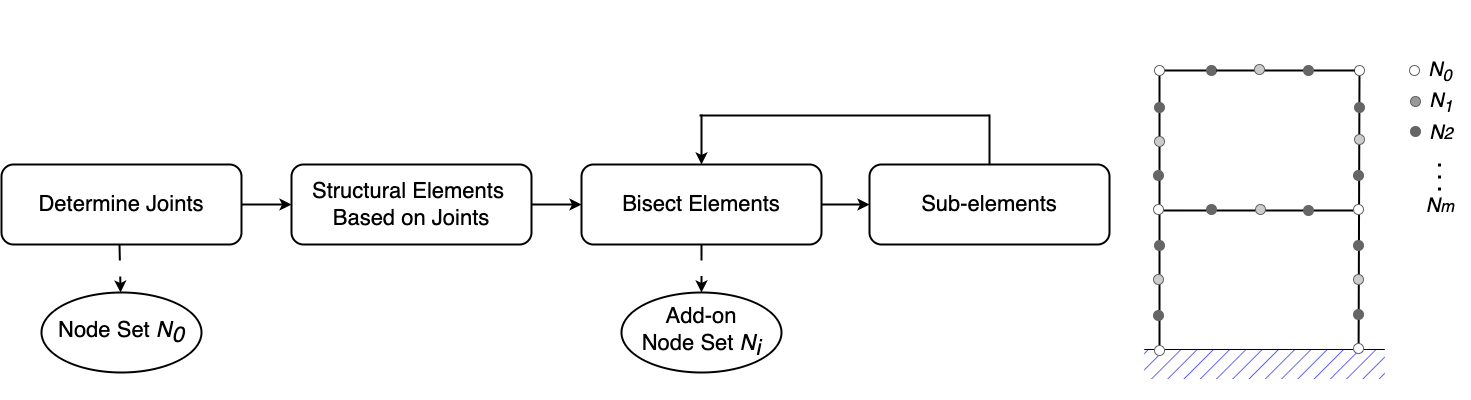}
}
\caption{The process of joint-first bisection. N0 is the set of nodes determined by joints. Ni is the set of the nodes extracted after the i-th time of bisecting the structural elements.\label{fig2}}
\end{figure*}

As illustrated in ~\autoref{fig2}, the graph is first built using our proposed joint-first bisection method. In its initial stage, each structural joint is defined as a node, and each structural element (e.g., beam or column) connecting two joints is represented as an edge, forming the original node set $N_0$. Here, the definition of a joint is not restricted to conventional beam-column intersections. It can also include locations where the cross-section changes abruptly.

To obtain a finer structural representation, the structural members are then progressively divided. In each refinement iteration, every structural element (i.e., edge) is bisected into two equal-length sub-elements, and the newly created intermediate points are added as additional nodes, transforming the graph to receive a refined node set $N_i$. This iterative process generates new edges and nodes after each bisection, as shown in the ~\autoref{fig2} by the transition from the original frame layout to a denser graph with beam and column sub-elements. After the final edge-bisection step, the total number of nodes is given by the sum of all node sets, namely $N_9$ + ... + $N_m$. Once the graph topology is established, geometric and mechanical attributes are assigned to the nodes and edges to define the input features used by the GNN.

As an initial step in exploring the use of GNNs for structural displacement prediction, only the frame joints are treated as nodes. Under this representation, the two-story frame contains six nodes and six structural members, allowing the model to capture both vertical and horizontal load-transfer mechanisms within the system. The fixed supports at the base are explicitly encoded in the node features, whereas the remaining joints are modeled as free to deform. To preserve the directionality of load transfer, each structural member is further represented as a pair of oppositely directed edges, with a directional flag included in the edge attributes.

% figure2

With respect to graph features, each node is assigned attributes that describe its geometry, boundary conditions, and loading state, as summarized in Table 1. These attributes include nodal coordinates, three binary indicators specifying whether the translational and rotational degrees of freedom are restrained, external loads ($F_x$, $F_y$, $M_z$), and an additional indicator identifying intermediate joints where beams and columns intersect. A scalar parameter, $\phi$, is further introduced to represent the strong-column–weak-beam (SCWB) requirement (see Table 3), thereby informing the model of the structural hierarchy. The edges are enriched with both geometric and mechanical attributes, as listed in Table 2. Geometric attributes include member length and orientation, expressed through the cosine and sine of the member angle. Mechanical attributes include axial stiffness (EA), flexural stiffness (EI), plastic section modulus (Z), and plastic moment capacity ($M_p = F_y \cdot Z$). In addition, a Boolean indicator is used to distinguish columns from beams. Together, these feature encodings enable the GNN to learn the relationship between member properties and the resulting global displacement distribution.

% table1
\begin{table*}[!h]
\caption{Node feature vector (per node, 10 dimensions)\label{tab:node_features}}
\begin{tabular*}{\textwidth}{@{\extracolsep\fill}clllll@{}}
\toprule
\textbf{Index} & \textbf{Name} & \textbf{Key} & \textbf{Type} & \textbf{Unit} & \textbf{Description} \\
\midrule
1 & Node x-coordinate   & x       & float   & m          & From NODE\_COORDS \\
2 & Node y-coordinate   & y       & float   & m          & From NODE\_COORDS \\
3 & BC flag $U_X$       & bc\_ux  & \{0,1\} & --         & 1 = fixed, 0 = free \\
4 & BC flag $U_Y$       & bc\_uy  & \{0,1\} & --         & 1 = fixed, 0 = free \\
5 & BC flag $R_Z$       & bc\_rz  & \{0,1\} & --         & 1 = fixed, 0 = free \\
6 & Nodal force $F_X$   & fx      & float   & N          & Applied to nodes 2 and 3 \\
7 & Nodal force $F_Y$   & fy      & float   & N          & Always 0 \\
8 & Nodal moment $M_Z$  & mz      & float   & N$\cdot$m  & Always 0 \\
\bottomrule
\end{tabular*}
\end{table*}

% table2
\begin{table*}[!h]
\caption{Edge feature vector (per directed edge, 10 dimensions).\label{tab:edge_features}}
\begin{tabular*}{\textwidth}{@{\extracolsep\fill}clllll@{}}
\toprule
\textbf{Index} & \textbf{Name} & \textbf{Key} & \textbf{Type} & \textbf{Unit} & \textbf{Description} \\
\midrule
1  & Length                   & L         & float    & m          & Distance between nodes \\
2  & Cosine                   & $\cos\theta$   & float    & --         & $dx/L$ \\
3  & Sine                     & $\sin\theta$   & float    & --         & $dy/L$ \\
4  & Axial stiffness          & EA        & float    & N          & $E \cdot A$ \\
5  & Bending stiffness        & EI        & float    & N$\cdot$m$^2$ & $E \cdot I$ \\
6  & Plastic section modulus  & Z         & float    & m$^3$      & Rectangular approx: $A \cdot h / 4$ \\
7  & Plastic moment           & $M_p$        & float    & N$\cdot$m  & $F_y \cdot Z$ \\
8  & Column flag              & is\_col   & \{0,1\}  & --         & 1 if vertical edge \\
9  & Beam flag                & is\_beam  & \{0,1\}  & --         & 1 if horizontal edge \\
10 & Direction                & dir       & \{+1,-1\} & --        & +1 forward, -1 backward \\
\bottomrule
\end{tabular*}
\end{table*}

Displacements ($U_x$, $U_y$) are expressed in millimeters, whereas rotations ($R_z$) are represented in degrees to mitigate scale imbalance and enhance numerical stability. All input and output features are normalized using the mean and standard deviation computed from the dataset, and these statistics are retained for use during inference to ensure consistency and reproducibility.

\subsection{Dataset Generation}
\begin{figure*}[t]
\centerline{
\includegraphics[width=0.3\textwidth]{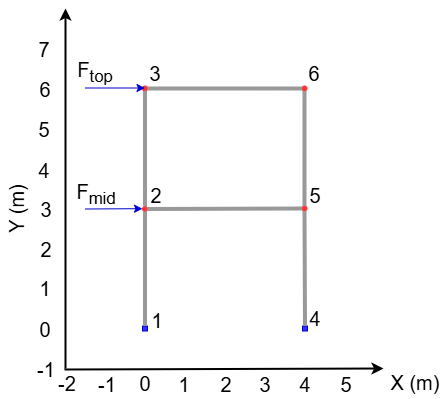}
}
\caption{The two-story frame structure investigated in this study was modeled in ANSYS to simulate its structural response under applied loading conditions.\label{fig3}}
\end{figure*}

All data are generated through ANSYS MAPDL simulations, with varying forces while keeping stiffness properties and boundary conditions constant. The specifications and parameters of the ANSYS model are shown in the ~\autoref{fig3} and ~\autoref{tab:dataset_parameters}. ~\autoref{tab:physical_properties} shows physical properties of columns and beams of the structure.

The load at the top and middle frames are selected to produce the linear and nonlinear deformations and both top and middle loads are randomly combined and then applied to the structure to produce the responses of the structure. The input and output values form the datasets that will be used for machine learning practices.

% figure3

% table3
\begin{table*}[!h]
\caption{ANSYS model operating conditions and global config.\label{tab:dataset_parameters}}
\begin{tabular*}{\textwidth}{@{\extracolsep\fill}llllll@{}}
\toprule
\textbf{Category} & \textbf{Key} & \textbf{Name} & \textbf{Description} & \textbf{Unit} & \textbf{Value/Range} \\
\midrule
Dataset  & N\_CASES   & Number of cases   & Total number of generated samples                     & -- & 500 \\
Loads    & F          & Force range       & Range of horizontal forces at mid/top nodes          & N  & 50{,}000--230{,}000 \\
Material & E          & Young's modulus   & Elastic modulus range of steel                       & Pa & $2.05 \times 10^{11}$ \\
Material & $F_Y$         & Yield strength    & Steel yield strength (for $M_p = Z \cdot F_y$)      & Pa & $3.45 \times 10^{8}$ \\
SCWB     & PHI\_SCWB  & SCWB factor      & Strong-column weak-beam factor ($\phi$)                       & -- & 1.2 \\
Geometry & story\_height & Story height   & Height of one story                                  & m  & 3.0 \\
Geometry & bay\_width & Bay width         & Span width                                           & m  & 4.0 \\
\bottomrule
\end{tabular*}
\end{table*}

% table4
\begin{table*}[!h]
\caption{Physical properties of structure's components.\label{tab:physical_properties}}
\begin{tabular*}{\textwidth}{@{\extracolsep\fill}lllllll@{}}
\toprule
\textbf{Member} & \textbf{$A$ (m$^2$)} & \textbf{$I$ (m$^4$)} & \textbf{$b$ (m)} & \textbf{$h$ (m)} & \textbf{$Z$ (m$^3$)} & \textbf{$M_p$ (N$\cdot$m)} \\
\midrule
Column & 0.08 & $3.5 \times 10^{-4}$ & 0.3491 & 0.2291 & $4.5826 \times 10^{-3}$ & $1.58 \times 10^{6}$ \\
Beam   & 0.04 & $2.5 \times 10^{-4}$ & 0.1461 & 0.2739 & $2.7386 \times 10^{-3}$ & $9.45 \times 10^{5}$ \\
\bottomrule
\end{tabular*}
\end{table*}

To determine whether the frame reaches the plastic stage, the lateral load level corresponding to first yield was estimated. This load threshold was then used to distinguish the linear and nonlinear datasets and to determine their proportions in the overall database.

Q235 is a commonly used structural carbon steel with elastic modulus \(E\), Poisson's ratio \(\nu\), and yield strength \(F_y\). In the stress--strain relationship, the material initially exhibits linear elastic behavior. Once the stress reaches the yield point, it enters the plastic stage, during which the stress increases with strain in a nonlinear manner.

The stress--strain curve in ANSYS was defined using a multilinear isotropic hardening (MISO) material model to simulate the elastic--plastic response of Q235 steel. Accordingly, the generated dataset contains both linear and nonlinear behavioral characteristics.

To estimate the first-yield load level of the frame structure under lateral loading, the Portal Method was adopted for approximate evaluation ~\cite{hanson_2020_structural}. This method simplifies the frame into an equivalent beam--column system and assumes that the beams yield prior to the columns. Therefore, the plastic bending moment of the beam is taken as the yielding criterion. Based on the geometric and material parameters, the equivalent rectangular section height about the strong axis is determined by ~\autoref{eq:hsec} and the plastic section modulus is expressed as ~\autoref{eq:Z}.

\begin{equation}
h_{\mathrm{sec}} = \sqrt{\frac{12I}{A}},
\label{eq:hsec}
\end{equation}

\begin{equation}
Z = \frac{A h_{\mathrm{sec}}}{4}.
\label{eq:Z}
\end{equation}

The beam yield moment is then given by ~\autoref{eq:My} and, under the assumptions of the Portal Method, the story shear threshold may be approximated as ~\autoref{eq:Vy}.

\begin{equation}
M_y = Z F_y,
\label{eq:My}
\end{equation}

\begin{equation}
V_y \approx \frac{2 M_y}{h_{\mathrm{story}}}.
\label{eq:Vy}
\end{equation}

When the story shear reaches this value, the structure may be considered to enter the nonlinear stage for the first time. In this study, Q235 steel was used with yield strength \(F_y = 2.35 \times 10^8~\mathrm{Pa}\), beam cross-sectional area \(A = 0.04~\mathrm{m}^2\), moment of inertia \(I = 2.5 \times 10^{-4}~\mathrm{m}^4\), and story height \(h_{\mathrm{story}} = 3.0~\mathrm{m}\). The calculated equivalent section height is \(h_{\mathrm{sec}} = 0.2739~\mathrm{m}\), and the plastic section modulus is \(Z = 2.7386 \times 10^{-3}~\mathrm{m}^3\).

Accordingly, the beam yield moment $M_y$ is defined as ~\autoref{eq:My_val} and the corresponding story shear threshold is defined approximately in ~\autoref{eq:Vy_val}.

\begin{equation}
M_y = 6.4357 \times 10^5~\mathrm{N \cdot m},
\label{eq:My_val}
\end{equation}

\begin{equation}
V_y = 4.29 \times 10^5~\mathrm{N}.
\label{eq:Vy_val}
\end{equation}

Based on this estimation, when a lateral load is applied only at the top story, the beam first yields at approximately 
$F_{\mathrm{top}} = 4.29 \times 10^5~\mathrm{N}.$
When the applied load is limited to $F_{\mathrm{top}} \lesssim 2.57 \times 10^5~\mathrm{N}$, the structure generally remains in the elastic stage. Both the top and middle stories are loaded simultaneously until the structure gradually approaches yielding. The lateral forces at the yield point are above the value showing in \autoref{eq:combined_load}.

\begin{equation}
\left(F_{\mathrm{top}},\, F_{\mathrm{mid}} + F_{\mathrm{top}}\right) > 2.57 \times 10^5~\mathrm{N}.
\label{eq:combined_load}
\end{equation}

A proportional load-increment scan was further conducted from \(0\) to \(F_{\max} = 8.0 \times 10^5~\mathrm{N}\) in 12 steps. By evaluating the variation of the story shear ratio (see \autoref{eq:shear_ratio}), the approximate load step at which the structure first enters the nonlinear range can be identified.

\begin{equation}
\left(\frac{V_1}{V_y},\, \frac{V_2}{V_y}\right),
\label{eq:shear_ratio}
\end{equation}

The results indicate that at Step 4 (load factor \(= 0.333\)), the horizontal loads applied at the top and middle stories are both approximately and the corresponding story shear ratio is \(1.24\), indicating that the structure has reached and slightly exceeded the yield threshold at this stage (see \autoref{eq:f_top_mid}).

\begin{equation}
F_{\mathrm{top}} = F_{\mathrm{mid}} = 2.67 \times 10^5~\mathrm{N},
\label{eq:f_top_mid}
\end{equation}

The Portal Method provides a rapid estimate of the yield threshold: responses within approximately \(2.6 \times 10^5~\mathrm{N}\) may be regarded as lying in the linear range, whereas those exceeding approximately \(4.3 \times 10^5~\mathrm{N}\) are considered to enter the nonlinear stage. This estimate provides an important basis for determining the loading range in the subsequent finite element analysis using the MISO material model.

\begin{figure*}[t]
\centerline{
\includegraphics[width=0.87\textwidth]{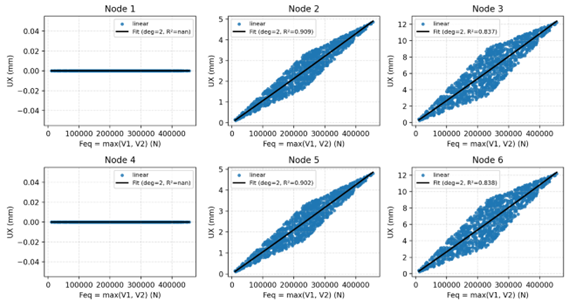}
}
\caption{The linear force-displacement relationship per node.\label{fig:linear_verification}}
\end{figure*}

\begin{figure*}[t]
\centerline{
\includegraphics[width=0.87\textwidth]{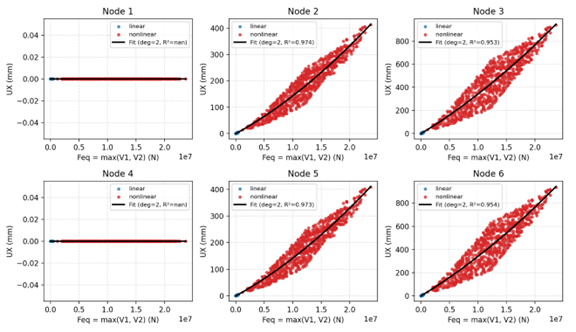}
}
\caption{The nonlinear force-displacement relationship per node.\label{fig:nonlinear_verification}}
\end{figure*}

~\autoref{fig:linear_verification} and ~\autoref{fig:nonlinear_verification} illustrate the verification of the linear and nonlinear datasets through the force--displacement response, represented by \(U_X\), at each node. As shown in ~\autoref{fig:nonlinear_verification}, the black line exhibits a nonlinear trend.

% figure4

% figure5

\subsection{Implementation}
Our proposed network is based on an edge-conditioned message-passing scheme. First, the input feature vector of each node is projected into a hidden space. As shown in \autoref{eq:hidden_space}, \(x_i\) denotes the feature vector of node \(i\), \(W_{\mathrm{in}}\) is the input weight matrix, and \(h_i^{(0)}\) is the initial hidden representation of node \(i\).

\begin{equation}
h_i^{(0)} = W_{\mathrm{in}} x_i
\label{eq:hidden_space}
\end{equation}

At each message-passing layer \(k\), the edge-conditioned message sent from node \(j\) to node \(i\), denoted by \(m_{ij}^{(k)}\), is computed as shown in \autoref{eq:layer_edge_pass}.

\begin{equation}
m_{ij}^{(k)} = W(e_{ij})\, h_j^{(k)}
\label{eq:layer_edge_pass}
\end{equation}

The hidden representation of each node is then updated by aggregating messages from its neighboring nodes using mean aggregation, as given in \autoref{eq:node_update}. After the final message-passing layer, the node-level prediction \(\hat{y}_i\) is obtained through \autoref{eq:final_node_prediction}. For each node \(i\), the predicted output variables are \(U_X\), \(U_Y\), and \(R_Z\).

\begin{equation}
h_i^{(k+1)} = \mathrm{ReLU}\!\left( \operatorname{AGG}_{j \in \mathcal{N}(i)} m_{ij}^{(k)} \right)
\label{eq:node_update}
\end{equation}

\begin{equation}
\hat{y}_i = W_{\mathrm{out}} h_i^{(L)}
\label{eq:final_node_prediction}
\end{equation}

The loss function of the GNN model, shown in \autoref{eq:loss}, is designed to incorporate physical constraints. For free nodes, the predicted responses are compared with the ground-truth values using the mean squared error. For fixed nodes, an additional penalty is introduced when the predicted responses deviate from zero. This dual-objective formulation enables the network to satisfy boundary conditions while accurately learning the displacement behavior of deformable nodes. The total loss is defined as

\begin{equation}
\mathrm{Loss}
=
\frac{1}{|F|}\sum_{i \in F}\left\lVert y_i-\hat{y}_i \right\rVert^2
+\lambda \cdot \frac{1}{|C|}\sum_{i \in C}\left\lVert y_i-0 \right\rVert^2
\label{eq:loss}
\end{equation}

where \(\lambda\) denotes the penalty weight applied to fixed nodes, \(\hat{y}_i\) denotes the predicted displacement and rotation vector at node \(i\) (i.e., \(U_X\), \(U_Y\), and \(R_Z\)), and \(y_i\) denotes the corresponding ground-truth response vector.

The network is implemented using the \texttt{NNConv} layer in PyTorch Geometric. In this architecture, each edge feature vector is mapped to a weight matrix by a small multilayer perceptron, referred to as EdgeNet, which dynamically modulates message passing according to member stiffness and geometry. The model starts with a linear projection of node features into a latent space, followed by three successive \texttt{NNConv} layers with ReLU activation. At each layer, node embeddings are updated by aggregating messages from neighboring nodes, weighted by the corresponding edge-conditioned filters. Finally, a multilayer perceptron maps the learned embeddings to the nodal displacement outputs, namely \(U_X\), \(U_Y\), and \(R_Z\). Mean aggregation is adopted because it provides stable performance for the relatively small and regular graph topology considered in this study.

\subsection{Model Comparison}
To evaluate the performance of the proposed GNN, a conventional Neural Network (NN) model is developed for comparison. Both models were trained on the same dataset generated from ANSYS MAPDL simulations, ensuring consistency in training conditions.

In the NN model, the input features are two external forces applied at the mid and top nodes and predict corresponding nodal displacements and rotations. Unlike the GNN, the NN does not encode structural topology or member properties; therefore, the NN model is solely trained on force-response relations. This comparison lets us investigate the pros and cons of incorporating structural encoding over a graph and see it is better than a purely data-driven force–response mapping.

\section{Evaluation}\label{sec4}
All machine learning experiments, including the training of both GNN and NN models, were conducted using an 85/15 split of the dataset for training and testing, respectively. The models were trained using the Adam optimizer with a learning rate of 0.001 and a batch size of 32 for up to 100 epochs.

% table5
\begin{table*}[!h]
\caption{Comparison between predicted and actual displacements. ($F_mid$=200KN, $F_top$=150KN).\label{tab:predicted_actual}}
\begin{tabular*}{\textwidth}{@{\extracolsep\fill}cllllll@{}}
\toprule
& \multicolumn{3}{@{}l}{\textbf{Predicted}} & \multicolumn{3}{@{}l}{\textbf{Actual}} \\
\cmidrule{2-4}\cmidrule{5-7}
\textbf{Node} & \textbf{$U_X$ (mm)} & \textbf{$U_Y$ (mm)} & \textbf{$R_Z$ (deg)} & \textbf{$U_X$ (mm)} & \textbf{$U_Y$ (mm)} & \textbf{$R_Z$ (deg)} \\
\midrule
1 & 0.000 & 0.000   & 0.000000    & 0    & 0 & 0 \\
2 & 3.645 & 0.034   & $-0.117794$ & 3.64 & 0 & 0 \\
3 & 8.992 & 0.048   & 0.078234    & 8.99 & 0 & 0 \\
4 & 0.000 & 0.000   & 0.000000    & 0    & 0 & 0 \\
5 & 3.598 & $-0.034$ & $-0.116543$ & 3.59 & 0 & 0 \\
6 & 8.954 & $-0.048$ & 0.079347    & 8.96 & 0 & 0 \\
\bottomrule
\end{tabular*}
\end{table*}

% table6
\begin{table*}[!h]
\caption{Comparison between predicted and actual displacements ($F_{\mathrm{mid}} = 123.5$ kN, $F_{\mathrm{top}} = 100$ kN).\label{tab:comparison_displacements_123_100}}
\begin{tabular*}{\textwidth}{@{\extracolsep\fill}cllllll@{}}
\toprule
& \multicolumn{3}{@{}l}{\textbf{Predicted}} & \multicolumn{3}{@{}l}{\textbf{Actual}} \\
\cmidrule{2-4}\cmidrule{5-7}
\textbf{Node} & \textbf{$U_X$ (mm)} & \textbf{$U_Y$ (mm)} & \textbf{$R_Z$ (deg)} & \textbf{$U_X$ (mm)} & \textbf{$U_Y$ (mm)} & \textbf{$R_Z$ (deg)} \\
\midrule
1 & 0.000 & 0.000    & 0.000000    & 0    & 0 & 0 \\
2 & 2.371 & 0.020    & $-0.074482$ & 2.35 & 0 & 0 \\
3 & 5.879 & 0.031    & 0.051957    & 5.85 & 0 & 0 \\
4 & 0.000 & 0.000    & 0.000000    & 0    & 0 & 0 \\
5 & 2.348 & $-0.020$ & $-0.075036$ & 2.33 & 0 & 0 \\
6 & 5.865 & $-0.032$ & 0.052701    & 5.83 & 0 & 0 \\
\bottomrule
\end{tabular*}
\end{table*}

% table7
\begin{table*}[!h]
\caption{Comparison between predicted and actual displacements ($F_{\mathrm{mid}} = 200$ kN, $F_{\mathrm{top}} = 200$ kN).\label{tab:comparison_displacements_200_200}}
\begin{tabular*}{\textwidth}{@{\extracolsep\fill}cllllll@{}}
\toprule
& \multicolumn{3}{@{}l}{\textbf{Predicted}} & \multicolumn{3}{@{}l}{\textbf{Actual}} \\
\cmidrule{2-4}\cmidrule{5-7}
\textbf{Node} & \textbf{$U_X$ (mm)} & \textbf{$U_Y$ (mm)} & \textbf{$R_Z$ (deg)} & \textbf{$U_X$ (mm)} & \textbf{$U_Y$ (mm)} & \textbf{$R_Z$ (deg)} \\
\midrule
1 & 0.000 & 0.000    & 0.000000    & 0     & 0 & 0 \\
2 & 4.371 & 0.042    & $-0.142424$ & 4.37  & 0 & 0 \\
3 & 11.011 & 0.080   & 0.099864    & 11.02 & 0 & 0 \\
4 & 0.000 & 0.000    & 0.000000    & 0     & 0 & 0 \\
5 & 4.323 & $-0.042$ & $-0.141015$ & 4.32  & 0 & 0 \\
6 & 10.946 & $-0.060$ & 0.100230   & 10.97 & 0 & 0 \\
\bottomrule
\end{tabular*}
\end{table*}

\subsection{Linear Deformation}
Three representative results under different load combinations are presented in \autoref{tab:predicted_actual}, \autoref{tab:comparison_displacements_123_100}, and \autoref{tab:comparison_displacements_200_200}. These loading cases were selected such that the entire structure remained within the linear deformation range. Among the three examples, only the case with 200 kN loads applied to both the top and middle beams produced horizontal displacements $U_X$  exceeding 10 mm at Nodes 3 and 6.

A tolerance-based criterion was adopted to evaluate prediction accuracy for two main reasons. First, exact numerical agreement is not always a reasonable requirement, as real-valued computations inherently involve approximation. Second, structural design and structural health monitoring commonly incorporate safety factors, implying that a bounded deviation from the reference value can still be acceptable in practical applications.

To quantitatively assess model performance, the same tolerance-based metrics were applied to both the GNN and NN models. For displacements $U_X$ , $U_Y$ , a prediction was considered accurate if it fell within 
±0.01mm of the reference value, whereas for rotations $R_Z$, a tolerance of 0.001 was used. Under these criteria, the GNN achieved an overall accuracy of 86.56\% on the test dataset. This result demonstrates the GNN model’s strong ability to generalize beyond the training samples while capturing the underlying structural response with high fidelity.

For comparison, the NN was trained and evaluated on the same dataset. Because the NN uses only the applied forces as inputs and does not explicitly encode the structural topology, its predictive performance is substantially lower. Using the same tolerance thresholds, the NN achieved an accuracy of 59.33\% on the test dataset, which is significantly below the accuracy of GNN. This comparison highlights the importance of incorporating graph-based structural information into the learning framework.

% table8
\begin{table*}[!h]
\caption{Comparison between predicted and actual displacements ($F_{\mathrm{mid}} = 800$ kN, $F_{\mathrm{top}} = 600$ kN).\label{tab:comparison_displacements_800_600}}
\begin{tabular*}{\textwidth}{@{\extracolsep\fill}cllllll@{}}
\toprule
& \multicolumn{3}{@{}l}{\textbf{Predicted}} & \multicolumn{3}{@{}l}{\textbf{Actual}} \\
\cmidrule{2-4}\cmidrule{5-7}
\textbf{Node} & \textbf{$U_X$ (mm)} & \textbf{$U_Y$ (mm)} & \textbf{$R_Z$ (deg)} & \textbf{$U_X$ (mm)} & \textbf{$U_Y$ (mm)} & \textbf{$R_Z$ (deg)} \\
\midrule
1 & 0.000  & 0.000    & 0.000000    & 0     & 0 & 0 \\
2 & 15.399 & 0.084    & $-0.492770$ & 16.03 & 0 & 0 \\
3 & 40.780 & $-0.004$ & $-0.285593$ & 41.15 & 0 & 0 \\
4 & 0.000  & 0.000    & 0.000000    & 0     & 0 & 0 \\
5 & 15.376 & $-0.156$ & $-0.447447$ & 15.88 & 0 & 0 \\
6 & 40.132 & $-0.249$ & $-0.320929$ & 40.96 & 0 & 0 \\
\bottomrule
\end{tabular*}
\end{table*}

%table9
\begin{table*}[!h]
\caption{Comparison between predicted and actual displacements ($F_{\mathrm{mid}} = 1000$ kN, $F_{\mathrm{top}} = 1000$ kN).\label{tab:comparison_displacements_1000_1000}}
\begin{tabular*}{\textwidth}{@{\extracolsep\fill}cllllll@{}}
\toprule
& \multicolumn{3}{@{}l}{\textbf{Predicted}} & \multicolumn{3}{@{}l}{\textbf{Actual}} \\
\cmidrule{2-4}\cmidrule{5-7}
\textbf{Node} & \textbf{$U_X$ (mm)} & \textbf{$U_Y$ (mm)} & \textbf{$R_Z$ (deg)} & \textbf{$U_X$ (mm)} & \textbf{$U_Y$ (mm)} & \textbf{$R_Z$ (deg)} \\
\midrule
1 & 0.000  & 0.000    & 0.000000    & 0     & 0 & 0 \\
2 & 21.573 & 0.118    & $-0.685229$ & 21.85 & 0 & 0 \\
3 & 49.160 & 0.003    & $-0.438357$ & 55.09 & 0 & 0 \\
4 & 0.000  & 0.000    & 0.000000    & 0     & 0 & 0 \\
5 & 21.285 & $-0.286$ & $-0.663005$ & 21.61 & 0 & 0 \\
6 & 49.202 & $-0.431$ & $-0.457481$ & 54.84 & 0 & 0 \\
\bottomrule
\end{tabular*}
\end{table*}

%table10
\begin{table*}[!h]
\caption{Each node's accuracy after training on GNN.\label{tab:node_accuracy_gnn}}
\begin{tabular*}{\textwidth}{@{\extracolsep\fill}lll@{}}
\toprule
\textbf{Node} & \textbf{Training Accuracy} & \textbf{Testing Accuracy} \\
\midrule
1 & 100.00\% & 100.00\% \\
2 & 92.87\%  & 92.67\%  \\
3 & 81.72\%  & 82.50\%  \\
4 & 100.00\% & 100.00\% \\
5 & 96.13\%  & 96.08\%  \\
6 & 80.5\%   & 80.01\%  \\
\bottomrule
\end{tabular*}
\end{table*}

%table11
\begin{table*}[!h]
\caption{GNN results.\label{tab:gnn_results}}
\begin{tabular*}{\textwidth}{@{\extracolsep\fill}lllllll@{}}
\toprule
\textbf{Dataset} & \textbf{Phase} & \textbf{Number of Datapoints} & \textbf{Overall Accuracy} & \textbf{$U_X$ Accuracy} & \textbf{$U_Y$ Accuracy} & \textbf{$R_Z$ Accuracy} \\
\midrule
Testing  & Overall   & 2000 & 91.87\% & 80.25\% & 100.00\% & 95.35\% \\
Testing  & Linear    & 1200 & 89.19\% & 75.32\% & 100.00\% & 92.25\% \\
Testing  & Nonlinear & 800  & 95.88\% & 87.65\% & 100.00\% & 100.00\% \\
Training & Overall   & 400  & 92.01\% & 81.00\% & 100.00\% & 95.04\% \\
Training & Linear    & 240  & 89.21\% & 75.86\% & 100.00\% & 91.77\% \\
Training & Nonlinear & 160  & 96.26\% & 88.78\% & 100.00\% & 100.00\% \\
\bottomrule
\end{tabular*}
\end{table*}

%table12
\begin{table*}[!h]
\caption{NN results.\label{tab:nn_results}}
\begin{tabular*}{\textwidth}{@{\extracolsep\fill}lllllll@{}}
\toprule
\textbf{Dataset} & \textbf{Phase} & \textbf{Number of Datapoints} & \textbf{Overall Accuracy} & \textbf{$U_X$ Accuracy} & \textbf{$U_Y$ Accuracy} & \textbf{$R_Z$ Accuracy} \\
\midrule
Testing  & Overall   & 2000 & 46.38\% & 49.89\% & 79.10\% & 47.85\% \\
Testing  & Linear    & 1200 & 58.94\% & 49.89\% & 79.10\% & 47.85\% \\
Testing  & Nonlinear & 800  & 27.55\% & 33.33\% & 32.23\% & 17.08\% \\
Training & Overall   & 400  & 46.48\% & 49.93\% & 79.32\% & 47.58\% \\
Training & Linear    & 240  & 58.94\% & 49.93\% & 79.32\% & 47.58\% \\
Training & Nonlinear & 160  & 27.81\% & 33.33\% & 32.81\% & 17.30\% \\
\bottomrule
\end{tabular*}
\end{table*}

\subsection{Nonlinear Deformation}
Two representative GNN prediction results under different load combinations are presented in \autoref{tab:comparison_displacements_800_600} and \autoref{tab:comparison_displacements_1000_1000}. These loading cases were selected such that the entire structure exhibited nonlinear deformation.

As shown in \autoref{tab:comparison_displacements_1000_1000}, the discrepancy between the predicted and actual horizontal displacements, $U_X$, increases substantially at Nodes 3 and 6. This trend suggests that these nodes are approaching the plastic deformation range, in which the structural response exhibits stronger nonlinearity and is consequently more difficult to predict with high accuracy. In comparison with the results obtained in the linear deformation range, this observation indicates a deterioration in model performance under nonlinear conditions. Nevertheless, despite Nodes 3 and 6 showing the lowest prediction accuracies among the six nodes, both still maintain accuracies exceeding 80\% (see \autoref{tab:node_accuracy_gnn}), demonstrating that the GNN is capable of producing reasonably reliable predictions for the two-story frame even in the nonlinear regime. Furthermore, based on the earlier comparison in the linear case, it is reasonable to infer that the NN model would exhibit even poorer predictive performance under the same nonlinear deformation conditions.

\subsection{Overall Deformation}
When comparing the training results between GNN and NN models in entire range (i.e, both linear and nonlinear phases), we set a tolerance range for $U_X$ , $U_Y$ , and $R_Z$ . This includes the consideration of widen the tolerance in the nonlinear zone. In the linear zone, $U_X$  and $U_Y$  allow ±0.2 mm displacement difference and $R_Z$  allows ±0.05° rotation difference. Due to consideration of the large deformation, we tolerate larger error range. In the nonlinear zone, $U_X$  and $U_Y$  allow ±1.0 mm displacement difference, and $R_Z$  allows ±1.0° rotation difference.

\autoref{tab:gnn_results} and \autoref{tab:nn_results}  show that the results demonstrate that the GNN provides both higher accuracy and better physical consistency compared to the NN, particularly in predicting nodal displacements and rotations under varying load conditions. We think that GNN’s performance is attributed by the edge-conditioned message passing mechanism, which adaptively updates nodal features according to member stiffness and geometric relationships. As a result, this enables nonlinear structural behavior to be more effectively represented within the graph framework. Because the GNN model achieved higher accuracy when it is trained solely on linear datasets under the same tolerance than those trained on mixed datasets, suggesting that further improvement is still required for highly nonlinear response prediction.

\section{Conclusion}\label{sec5}
This study demonstrates that our GNN constitutes a feasible and efficient surrogate for the FEM in structural displacement prediction. By explicitly encoding the topology of the structural frame, the GNN preserve the geometric and mechanical relationships among members and nodes, enabling them to capture structural responses more effectively than the NN model because the NN model exhibited weaker results, particularly for rotational response ($R_Z$ ), underscoring the importance of incorporating structural topology and stiffness information into the learning framework.

The superiority of the GNN became more pronounced in the nonlinear range, where yielding and large deformations occurred. While the NN showed a substantial reduction in predictive accuracy and failed to adequately represent post-yield behavior, the GNN maintained an high accuracy, indicating strong performance beyond the elastic regime. 

Overall, the results confirm that GNNs can incorporate structural mechanics into the learning process and achieve greater physical consistency, smoother prediction fields, and superior generalization under varying loading conditions than conventional NNs. In addition, once trained, the GNN enables rapid inference, highlighting its potential for real-time structural monitoring and early warning applications. These findings support the use of GNN model as efficient surrogate models for FEM. 

As this study represents a preliminary investigation into the application of Graph Neural Networks (GNNs) for structural displacement prediction, considerable opportunities remain for further development. In particular, the performance of the model should be examined for more complex structural configurations with larger numbers of nodes and members in order to evaluate its scalability and robustness. Moreover, the incorporation of multi-task learning techniques for the concurrent prediction of inter-story drift ratios, natural frequencies, and damage indices may further expand its potential in structural health monitoring. Finally, the inclusion of experimental data, in addition to numerical simulation results, would provide a more rigorous basis for validating the practical applicability and reliability of the proposed approach.

%Bibliography
\bibliographystyle{unsrt}  
\bibliography{references}

\end{document}